\documentclass[letterpaper]{article} 
\usepackage{aaai24}  
\usepackage{times}  
\usepackage{helvet}  
\usepackage{courier}  
\usepackage[hyphens]{url}  
\usepackage{graphicx} 
\usepackage{natbib}  
\usepackage{caption} 
\usepackage{appendix}
\usepackage{algorithm}
\usepackage{algorithmic}
\usepackage{newfloat}
\usepackage{listings}
\DeclareCaptionStyle{ruled}{labelfont=normalfont,labelsep=colon,strut=off} 
\floatstyle{ruled}
\newfloat{listing}{tb}{lst}{}
\floatname{listing}{Listing}
\usepackage{overpic}
\usepackage{subfig}
\usepackage{multirow}
\usepackage{booktabs}
\usepackage{subcaption}
\usepackage{xcolor}

\title{Scaling and Masking: A New Paradigm of Data Sampling for Image and Video Quality Assessment}
\author {
	Yongxu Liu\textsuperscript{\rm 1,\rm 2}\thanks{Corresponding author.},
	Yinghui Quan\textsuperscript{\rm 2,\rm 1},
	Guoyao Xiao\textsuperscript{\rm2,\rm1},
	Aobo Li\textsuperscript{\rm 3},
	Jinjian Wu\textsuperscript{\rm 3}
}
\affiliations {
    \textsuperscript{\rm 1}Hangzhou Institute of Technology, Xidian University, Hangzhou 311200, China\\
    \textsuperscript{\rm 2}School of Electronic Engineering, Xidian University, Xi'an 710071, China\\
    \textsuperscript{\rm 3}School of Artificial Intelligence, Xidian University, Xi'an 710071, China\\
    \{yongxu.liu, gyxiao\}@xidian.edu.cn, \{yhquan, jinjian.wu\}@mail.xidian.edu.cn, abli@stu.xidian.edu.cn
}

\usepackage{bibentry}

\begin{document}

\maketitle

\begin{abstract}
Quality assessment of images and videos emphasizes both local details and global semantics, whereas general data sampling methods (e.g., resizing, cropping or grid-based fragment) fail to catch them simultaneously.
To address the deficiency, current approaches have to adopt multi-branch models and take as input the multi-resolution data, which burdens the model complexity.
In this work, instead of stacking up models, a more elegant data sampling method~(named as SAMA, \underline{s}c\underline{a}ling and \underline{ma}sking) is explored, which compacts both the local and global content in a regular input size.
The basic idea is to scale the data into a pyramid first, and reduce the pyramid into a regular data dimension with a masking strategy.
Benefiting from the spatial and temporal redundancy in images and videos, the processed data maintains the multi-scale characteristics with a regular input size, thus can be processed by a single-branch model. 
We verify the sampling method in image and video quality assessment. 
Experiments show that our sampling method can improve the performance of current single-branch models significantly, and achieves competitive performance to the multi-branch models without extra model complexity. 
The source code will be available at \url{https://github.com/Sissuire/SAMA}.

\end{abstract}

\section{Introduction}

Image/video quality assessment~(I/VQA) is to quantify the perceptual quality/feeling of the given image/video data from the perspective of users~\cite{liu2019tmm, heke2022, bvqi2023, bvqiplus2023, discovqa2023}. 
The best choice of measuring perceptual quality is to conduct a subjective experiment and collect the mean opinion scores~(MOS) from a group of subjects~\cite{fang2022perceptual, jiang2022sisr}. However, the huge cost and inefficiency of the subjective quality assessment cannot satisfy most demands in reality, thus pushing the development of I/VQA methods.
As user-generated content~(e.g., selfies, short videos) is dramatically overflowing in everyone's mobile phone, every platform~(such as TikTok and Facebook), and the whole internet, I/VQA serves as the criterion for quality monitoring, data screening or preference recommendation, and is becoming more active in the society of multimedia and computer vision~\cite{metaiqa2020, clipiqa2023, mdvqa2023}. 

\begin{figure}[h]  
\centering
\subfloat[Scaling/cropping]{\includegraphics[width=.45\linewidth]{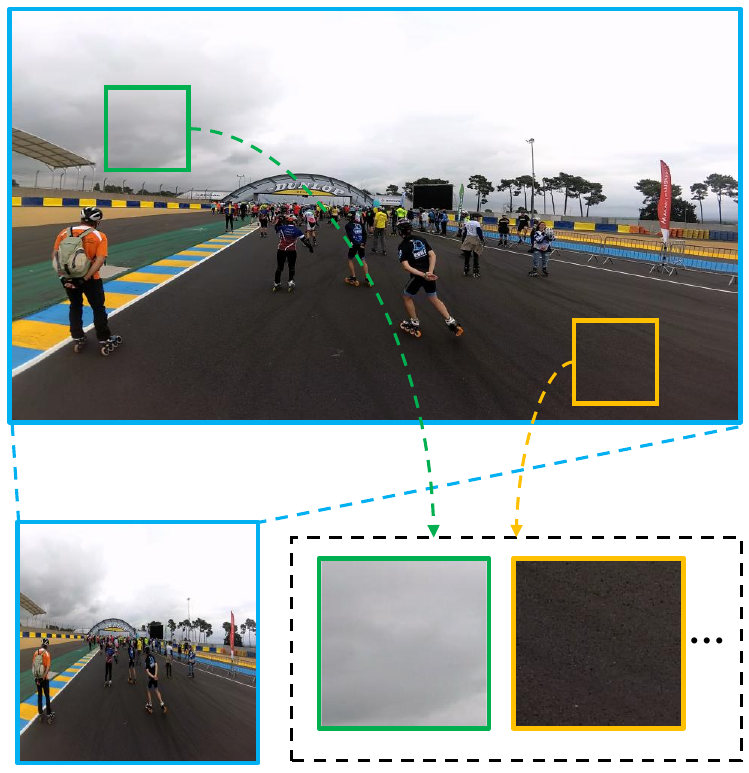}}\quad
\subfloat[Grid-based fragment]{\includegraphics[width=.45\linewidth]{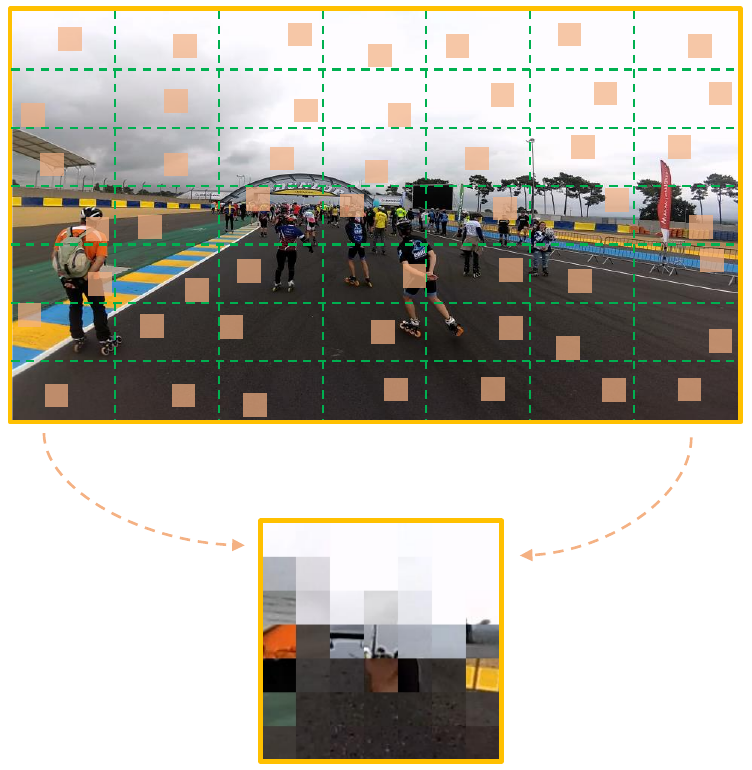}}\\
\subfloat[The proposed sampling method]{\includegraphics[width=0.9\linewidth]{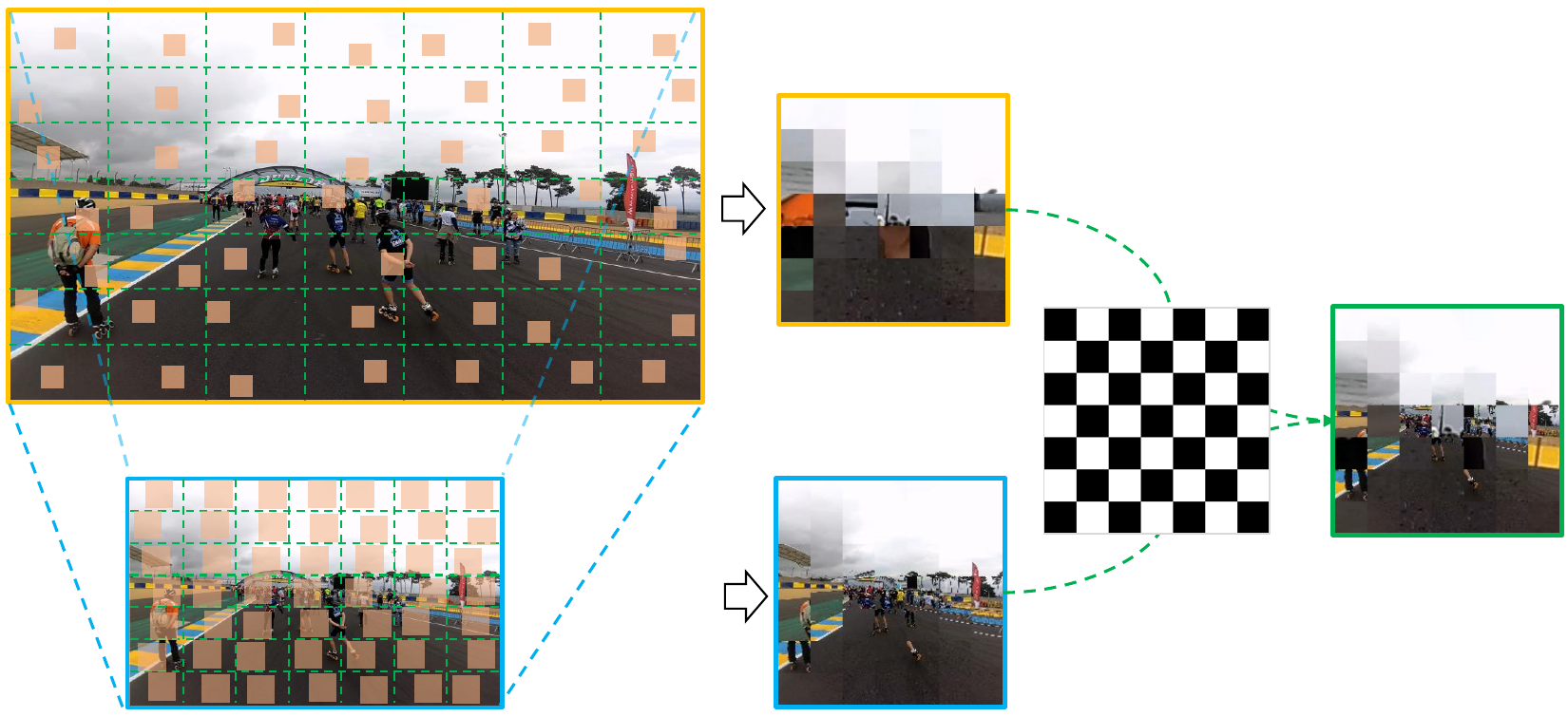}} 
\caption{An illustration of data sampling methods in quality assessment. Scaling would cause detail loss, while cropping might harm global perception. The proposed method scales the data into a pyramid and masks the pyramid based on data redundancy. The resulting data holds the multi-scale nature with a regular input size.}
\label{fig:intro}
\end{figure}

The intrinsic nature of quality assessment implies that the sampled data prepared for the quality assessment model should well preserve both local details and global appearance, which is not trivial for the fixed input size in deep learning-based models. Before inputting the data into the quality assessment model, a natural process is to interpolate the data into a fixed size~(e.g., $224\times224$). However, the scaling method would introduce the loss of local details, which is inappropriate for I/VQA. Previous work has suggested that preserving the higher resolution of the input data is critical~\cite{koniq}. As a consequence, cropping-based sampling is widely used in both IQA and VQA~\cite{p2p2020, patchvqa2021}. The concern is when facing an image with $1920\times1080$ or a higher resolution, the cropped patch might be too local to represent the general quality of the image.  
As shown in Fig.~\ref{fig:intro}(a), the cropped patches could be meaningless~(too flat). But more importantly, the local details cannot reflect the global semantics, which is also a key factor in I/VQA~\cite{sfa2018, vsfa2019}. 
Recently, FAST-VQA~\cite{fastvqa2022, fastervqa2023} extends the patch/cropping-based method and proposes a grid-based fragment sampling. Fragments are sampled at the raw resolution within each grid to preserve the local details~(as shown in Fig.~\ref{fig:intro}(b)). The SwinTransformer~\cite{swin2021, videoswin2022} is adopted to connect fragments and infer the overall semantics as well as perceptual quality. Still, it is not trivial to put all the global perception onto the model learning with limited fragments, especially for VQA where the number of annotated data is much less than the recognition task.
 
A natural solution to address the local and global trade-off is to build multi-branch models, where one for the raw fragments to hold the local details, and the other for the scaled input to see the global semantics~\cite{rich2021, dover2023, zoomvqa2023}. However, The multi-branch schemes introduce extra model complexity.
 MUSIQ~\cite{musiq2021} transforms the input data into a multi-scale image representation, and feeds them all into the model. Although the single-branch model is adopted, the dimension of input data is increased due to the multi-scale sampling, leading to a quadratic growth of  the model complexity in attention operation. In summary, both of them introduce much more computational complexity. So, the question is, can it be more elegant and sacrifice less computation to reach a good balance of both local preservation and global perception?
 
In this work, we propose a new paradigm of data sampling for I/VQA, named as SAMA~(i.e., \underline{s}c\underline{a}ling and \underline{ma}sking), which is highly motivated by the previous grid-based fragment~\cite{fastvqa2022}, multi-scale sampling~\cite{musiq2021} and masking strategy~\cite{mae2022}. As shown in Fig.~\ref{fig:intro}(c), the core idea is to build a multi-scale pyramid first to hold the multi-granularity image representation from local to global, then mask the pyramid based on spatial/temporal redundancy~(corresponding to image/video) into a regular input size. This masking strategy has a similar tuning/rearranging function as Bayer filter~\cite{bayer2013} does in imaging sensors, filtering different scaled data in a given pattern. The resulting data is scale-interlaced and shares the same size as the regular sampled one, thus can be fed into a single-branch base model. 
With a baseline model, we verify our proposed sampling method in both image and video quality assessment databases. Further, a branch of relative scale embedding methods are also explored to examine the consistency.

The main contributions  can be summarized as follows:

\begin{itemize}
    \item A novel data sampling method based on scaling and masking is introduced, to address the paradox of multi-granularity perception with the regular dimension.
    \item A group of relative scale encoding methods are specifically explored for the manipulated data structure.
    \item The method is verified on both IQA and VQA databases, and experimental results are impressive with almost negligible computation burden. 
\end{itemize}

\section{Related Work}

In this section, we give a brief view of both IQA and VQA from the perspective of data sampling.
\subsubsection{Blind IQA} 
Blind IQA is to estimate the quality of images without reference information, which is broadly explored in recent years. 
In the early stage, hand-crafted methods~\cite{blind2012, nrvpd2019} are unaware of the input resolution, and the features are mostly extracted from the raw data. 
As deep models are becoming prevalent, data sampling seems to be more important. 
Various methods crop the image into multiple patches and label each patch with the overall MOS~\cite{kim2016fully, kim2018deep, hyperiqa2020, p2p2020}. The patch-based sampling causes an incomplete representation~(lack of global view), an inaccurate annotation~(the overall MOS for each patch), and a costing inference process where multiple patches would be sampled for the inference. To catch the global perception, some work would like to rescale the data first and then crop it into a fixed size~\cite{koniq, vcrnet2022, ieit2022, qpt2023}. However, the resize would definitely cause detail loss, which is not friendly to quality assessment. 
Some other work purely fed a multi-scale image representation~\cite{musiq2021}, which significantly increases the computational burden. A similar way is to build the multi-scale module~\cite{wu2020end} or simply multi-branch models to solve the different views of data~\cite{vtiqa2022, dacnn2022, reiqa2023}. Different from existing IQA methods, our proposed sampling method can hold the local and global views simultaneously. Compared with the single-branch models, our input data contains more complete representation, while compared with the multi-branch ones, the proposed method does not require extra model complexity.

\subsubsection{Blind VQA}
Similar to blind IQA above, blind VQA can also be classified into the raw resolution-based, rescaling-based, cropping-based, and multi-branch methods. The raw resolution is generally adopted in traditional methods~\cite{videoblind2014,tlvqm2019}. Some work adopts pretrained encoders to extract features so that the raw resolution can be preserved~\cite{vsfa2019}. However, the fixed encoders cannot be optimized for VQA task, which makes the features unrepresentative. PVQ~\cite{patchvqa2021} trains the model with video patches, but the same problem would be faced like the patch-based sampling in IQA. The recent FAST-VQA~\cite{fastvqa2022, fastervqa2023} proposes to sample grid-based fragments and inference the global perception via SwinTransformer. The method partially relieves the local and global trade-off, but burdens the model learning. When the raw resolution increases, the global inference is thought to be harder. Our method is greatly inspired by the fragment sampling, and moves a step further. We introduce a pyramid of fragments to preserve the local and global views, and propose the masking strategy to reduce the pyramid to the same input size as FAST-VQA.  
Recently, some multi-branch models are proposed, and they generally construct two streams for both the local and global perception~\cite{zoomvqa2023, dover2023, maxvqa2023}. The performance is promising, but the complexity is also increasing. 

\begin{figure*}[h]  
\centering
\includegraphics[width=0.9\linewidth]{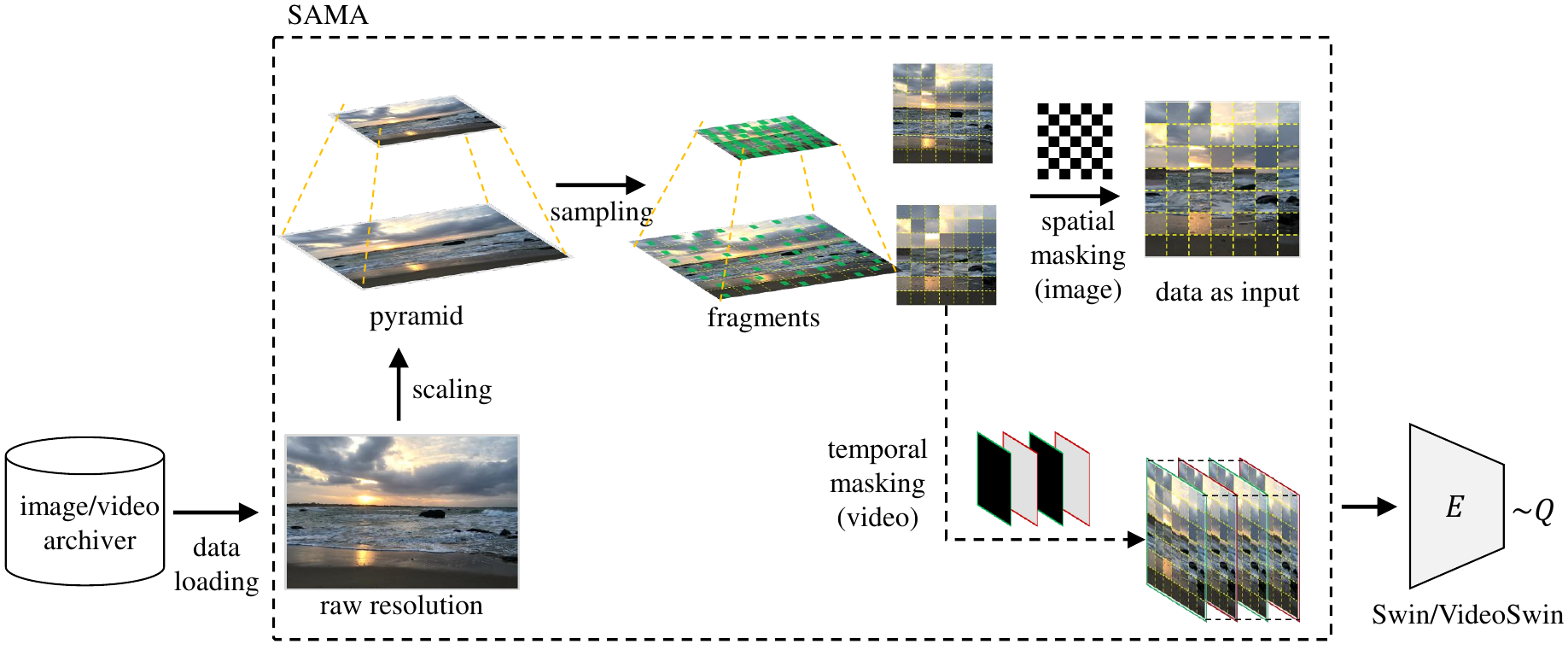}
\caption{The workflow of SAMA. Image or video data is first scaled into a multi-granularity pyramid via interpolation. Then fragments are sampled in each scale. Afterwards, spatial/temporal masking is constructed to tune the hierarchical fragments into a regular sampling size. The data after SAMA is fed into a base model for quality estimation.}
\label{fig:method}
\end{figure*}

\section{Proposed Method}

\subsection{Overall Architecture}
The goal of this work is to explore a more elegant data sampling method to achieve the multi-granularity perception in the layer of input data without extra dimension increase. 
With the data representation, a single-branch model could be easier to capture the local details and global semantics, and achieve similar performance as multi-branch models but with less complexity.
To guarantee the multi-granularity, the data would be scaled into the multi-scale pyramid at first. 
In order to achieve a regular input size, the dimension of the pyramid is expected to be reduced in an appropriate way. 

In this work, motivated by the content redundancy in image/video~\cite{ivc1997} and masking strategy~\cite{mae2022}, scales in the pyramid are masked and interlaced into a regular input size. 
The workflow is shown in Fig.~\ref{fig:method}.
Our proposed SAMA is based on fragment sampling~\cite{fastvqa2022}, and includes three key steps: scaling, sampling, and masking. After loading the raw resolution data, a multi-granularity pyramid is built via scaling. Then, the grid-based fragment sampling is adopted to extract local content for each scale, and forms a pyramid of fragments. Afterwards, we mask and tune the fragments in different scales based on spatial and temporal redundancy, and obtain the sampled data as model input. the sampled data is directly fed into a baseline model~\cite{fastvqa2022, swin2021, videoswin2022} to predict the perceptual quality. 

\subsection{Grid-based Fragment Sampling}
As SAMA is based on the fragment sampling~\cite{fastvqa2022}, we first give a short review in this subsection. 

Fragment sampling is explored initially for VQA, but the core idea is also suitable for IQA or other tasks. Given the data $x$~(image or video) with the raw resolution, fragment sampling first segments $x$ into $G_h\times G_w$ grids. Within each grid, a $f_h\times f_w$ patch at raw resolution is sampled. Afterwards, the data $x$ of any size could be sampled as $\hat{x}$ with a fixed size of $S=(G_h\cdot f_h)\times(G_w\cdot f_w)$. We omit the temporal dimension for simplicity here. In the work, the size of the grid is set as $7\times7$ and each patch is $32\times32$, thus $\hat{x}$ is with the size of $S=224\times224$. 
With the fragment sampling, the model input $\hat{x}$ preserves the raw details and a uniform distribution among the spatial content due to the grid partition. Besides, the discontinuity among near patches could be dissolved by Transformer-based architectures.

The fragment sampling preserves the local details of raw data, but puts the burden of global perception on model inference. When the resolution is increasing, the model might be hard to reason the global semantics from the sampled fragments. In this work, we construct a scale pyramid to relieve the dilemma and achieve a good trade-off of both local preservation and global perception.

\subsection{Scaling and Masking in Image}

\begin{figure}[t]  
\centering

\setlength{\abovecaptionskip}{0.15cm}
\subfloat[Window-based]{\includegraphics[width=.4\linewidth]{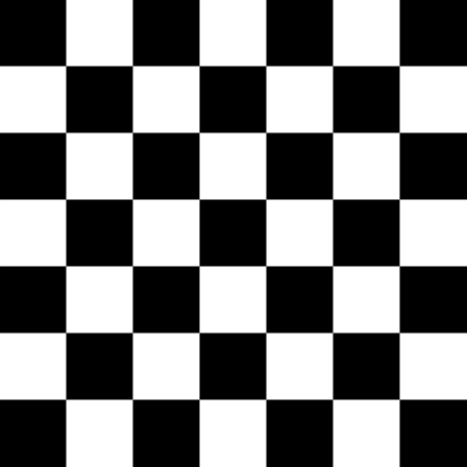}}\quad
\subfloat[Patch-based]{\includegraphics[width=.4\linewidth]{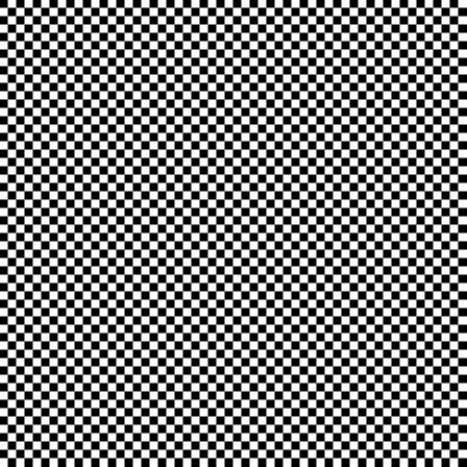}}\\
\subfloat[Progressive]{\includegraphics[width=.3\linewidth]{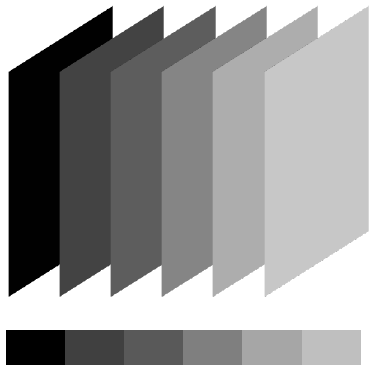}}\quad
\subfloat[Choppy]{\includegraphics[width=0.3\linewidth]{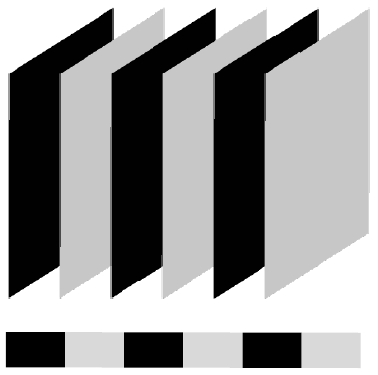}} \quad
\subfloat[Mixed]{\includegraphics[width=.3\linewidth]{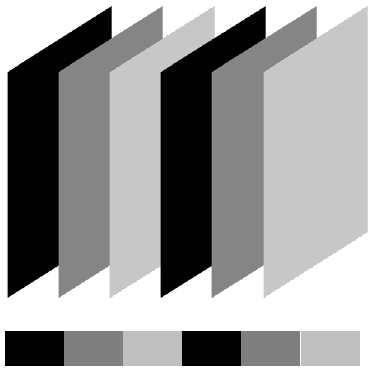}}
\caption{The illustration of spatial and temporal masks. (a) and (b) are spatial masks for images, and the last three are temporal masks for videos. Different intensities indicate the different scales.}
\label{fig:mask}

\end{figure}

For simplicity, the proposed SAMA is first clarified for image, then it would be extended into video in the following subsection.
Considering the great burden of model learning in fragment sampling, we are motivated by MUSIQ~(\cite{musiq2021}), an IQA method with multi-scale image representation, and introduce the multi-granularity pyramid to hold more data information from local to global. 

Given the image data $x$, a pyramid $p_x$ is first built as 
\begin{equation}
p_x=\{x_0, x_1, \cdots, x_n\}, 
\end{equation}
where $x_0$ is the same as $x$, and $x_i~(i>0)$ is a downsampled version of $x$ through bilinear interpolation but still keeps the aspect ratio of the image. We denote $x_n$ as the image with the smallest resolution 
satisfying $\mathrm{min}(h_{x_n}, w_{x_n}) = \mathrm{min}(G_h\cdot f_h, G_w\cdot f_w)$. 
where $h_{x_n}, w_{x_n}$ is the height and width of $x_n$. In practice, we uniformly scale the image thus the resolution of $x_i$ in $p_x$ decreases linearly from $x_0$ to $x_n$. 

With the multi-granularity pyramid, the fragment sampling is adopted to extract content and converts the pyramid with various resolutions into the representation with a fixed size~(e.g., $224\times224$). Thus, a pyramid of fragments is constructed as  
\begin{equation}
\hat{p_x} = \{\hat{x_0}, \hat{x_1}, \cdots, \hat{x_n}\}, 
\end{equation}
where $\hat{x_i}$ is the fragment sampled from $x_i$ with the resolution of  $S$. Note that, although each $\hat{x_i}$ shares the same resolution, fragments among different $\hat{x_i}$ are sampled from different scales in the multi-granularity pyramid, thus the fragments $\hat{p_x}$ can be also seen as a pyramid. 

The pyramid of fragments dramatically increases the data size from $S$ to $(n+1)\times S$. Some methods directly fed the pyramid into the model, thus increasing the complexity linearly or quadratically. The previous study has shown that the image content is largely redundant and thus can be greatly compressed~\cite{ivc1997}. Further, the recent masked autoencoder also suggests that even if the image content is masked with a large ratio, the content could still be recognized~\cite{mae2022}. The work has greatly inspired us to introduce the masking strategy and reduce the dimension of pyramid to a regular input size of $S$.

In the process of mask designing, we also refer to the Bayer pattern in imaging sensors, where the R/G/B is staggered with a certain arrangement. In this work, we introduce a scale-interlaced masking pattern that tunes the pyramid and transforms it into a regular input. Fig.~\ref{fig:mask}(a) and (b) give the illustration of the spatial masks for images. For simplicity, we only adopt two scales here, thus $n=1$. In Fig.~\ref{fig:mask}, we use different intensities of luminance to indicate the scales. Thus, in (a) and (b), the black positions correspond to the fragments sampled from $x_0$ while the white indicates those from $x_1$. When $n=1$, the data can be formulated as 
\begin{equation}
\tilde{x}=M\cdot \hat{x_0} + (1 - M) \cdot \hat{x_1},
\end{equation}
where $M$ is the spatial binary mask, in which 1/0 indicates the raw/scaled fragments.

In Fig.~\ref{fig:mask}, we provide two types of spatial masks. As our work is built on the fragment sampling, we choose the same baseline model, SwinTransformer~\cite{swin2021}, for quality assessment. According to the model architecture, we construct a window-based mask and a patch-based mask for images. For the window-based mask, the size of each black/white region is equal to the window size~(i.e., $32\times 32$) in SwinTransformer. Similarly, the size of each black/white region in the patch-based mask is the same as the embedding patch~(i.e., $4\times4$). Note that the mask can be extended to adapt to more scales. A very familiar way is to imitate Bayer RGGB in imaging sensors or the diamond arrangement in display that staggers multiple scales with a regular size.

\begin{figure}[t]  
\centering

\setlength{\abovecaptionskip}{0.15cm}
\subfloat[Raw data $x$]{\includegraphics[width=.7\linewidth]{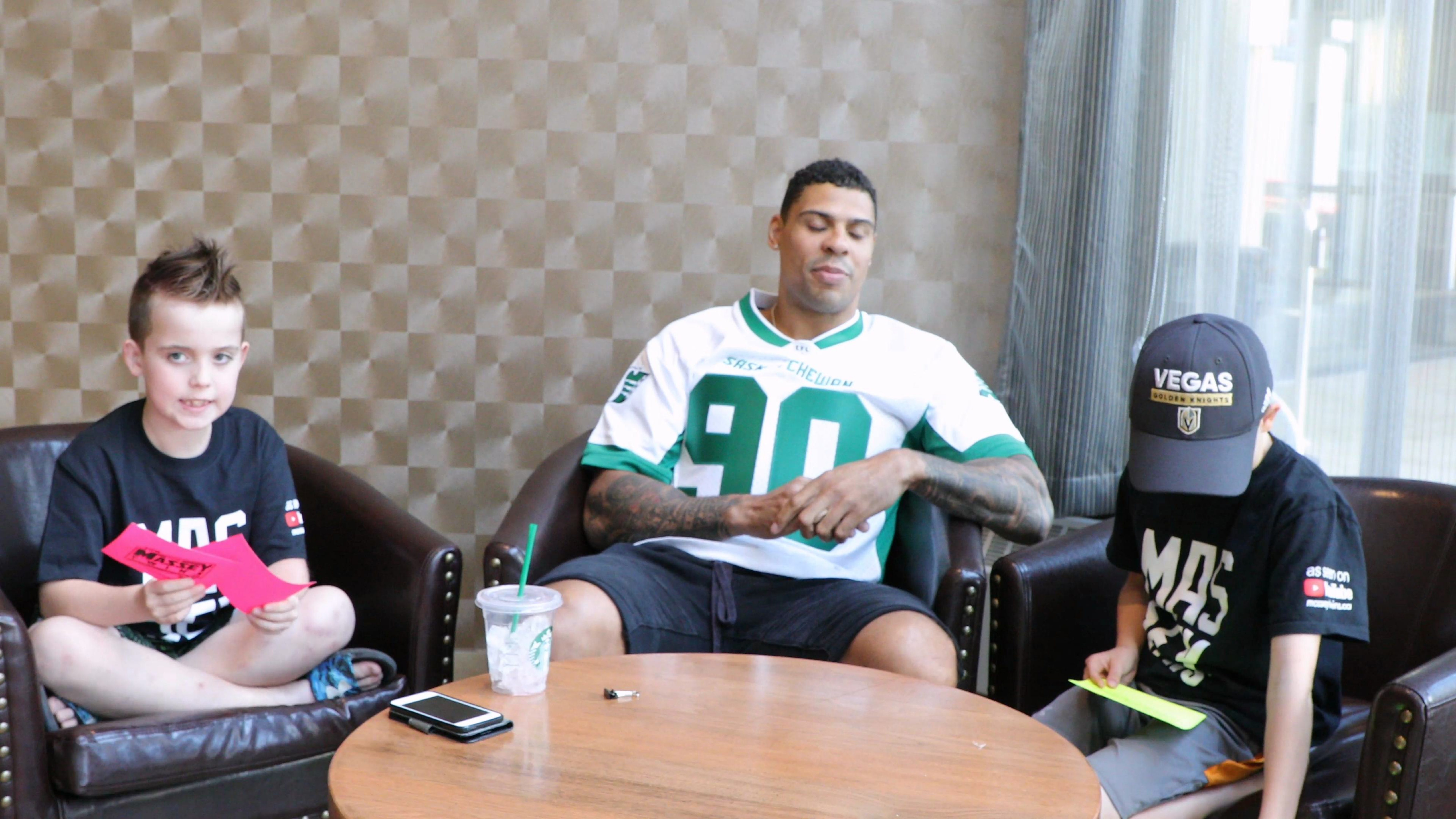}}\\
\subfloat[Fragments on $x_0$]{\includegraphics[width=.31\linewidth]{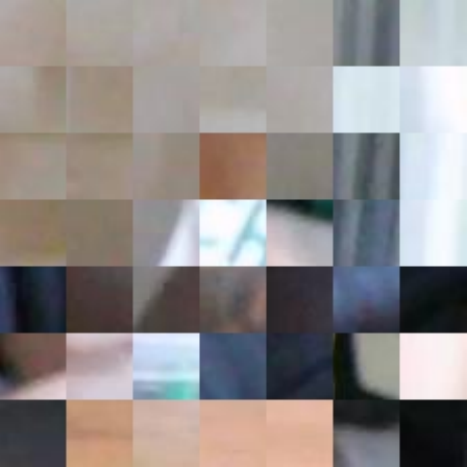}}\hfill
\subfloat[Fragments on $x_n$]{\includegraphics[width=.31\linewidth]{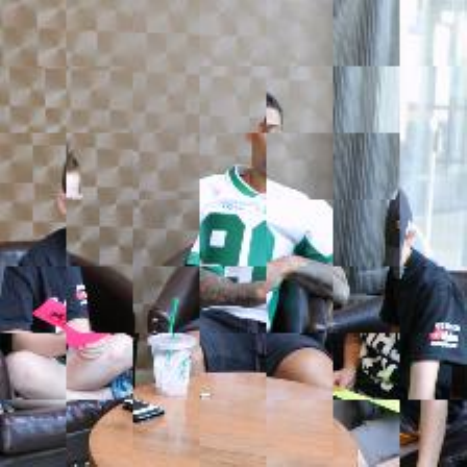}}\hfill
\subfloat[SAMA result]{\includegraphics[width=0.31\linewidth]{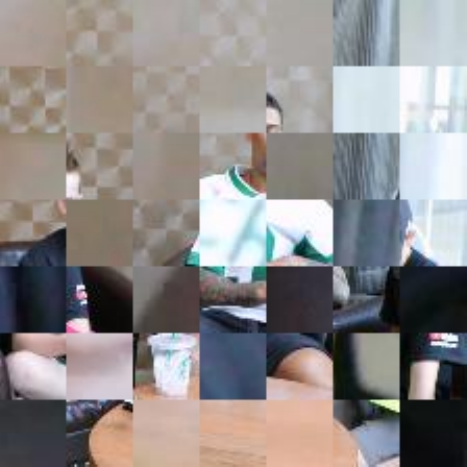}} 
\caption{An example of sampling result for image.}
\label{fig:example}
\end{figure}

Fig.~\ref{fig:example} gives an example where (b) shows the fragments sampled from raw resolution, and (c) shows the ones from scaled data. It shows that fragments sampled from $x_0$ focus more on the local details, and are hard to infer the global semantics. The fragments sampled from $x_n$ contain more global information, but miss the local distortions. With the masking strategy, the proposed SAMA masks and rearranges the multi-scale data into a regular shape, holding both local details and global semantics in a regular data shape.

\subsection{SAMA for Video}

When it comes to videos, the scaling and sampling are similar. The main difference is in the masking. Video data preserves not only the spatial correlation, but rich temporal redundancy. Thus, besides the spatial masks, we can explore more temporal masks for video. 

Given a video data $v$ containing $T$ frames, the scaling procedure transforms $v$ into the scale space  $p_v$, and the sampling extracts content from $p_v$ to generate a pyramid of fragments $\hat{p_v}$. In the process of masking, we introduce another three temporal masks based on temporal redundancy. As illustrated in Fig.~\ref{fig:mask}(c)$\sim$(e), a progressive mask indicates that, along the time dimension, the fragments are selected from the finest to the coarsest gradually. Considering the pyramid of fragments $\hat{p_v}$, where each scale actually contains $T$ frames since fragment sampling would not reduce the temporal dimension. What progressive mask does is to select the first two frames of fragments with raw resolution as $\tilde{v}_0$, and select the next two frames of fragments with a coarser resolution from the pyramid as $\tilde{v}_1$, until the coarsest. Thus, we can obtain the new data formed as
\begin{equation}
\tilde{v} = \{\tilde{v}_0, \tilde{v}_1, \cdots, \tilde{v}_{\frac{T}{2}-1}\},
\end{equation}
where each $\tilde{v}_i$ contains two frames of scaled fragments from the pyramid.
The two-frame setting results from the selection of our baseline model~(VideoSwin), where the temporal dimension of embedding patch is 2. As VideoSwin model takes as input 32 frames, thus the input data contains 16 scales of fragments, where the first two frames are the finest resolution, and the last two are the coarsest.

Besides the progressive masking, we also provide a choppy masking and a mixed one. As shown in Fig.~\ref{fig:mask}(d), the data form contains only two scales and they are interlaced in the temporal dimension. Still, each scale also contains two frames of fragments. For a 32-frame input data, the first two frames are the finest scale, and the next two frames are the coarsest. The arrangement would repeat 8 times to get a 32-frame input data. And for the mixed mask, the first 16 frames follow a progressive procedure, and the arrangement would repeat it to obtain 32 frames.

\subsection{Model Architecture and Implementation Details}

Our work serves as the data sampling method. Afterwards, the data is fed into a baseline model.
Our implementation is based on FAST-VQA~\cite{fastvqa2022}, and adopts the SwinTransformer/VideoSwin~\cite{swin2021, videoswin2022} for image/video tasks. For VQA, we simply substitute the original sampling method in FAST-VQA with SAMA, and keep all the other settings~(e.g., loss function, learning rate, batch size, etc.) as default. More details of the implementation can refer to the work~\cite{fastvqa2022} or our appendix.

Given the input size $S=H\times W$ for images, considering a $4\times 4$ patch embedding, the output of the baseline model, denoted as $z=f(\tilde{x})$, is $\frac{H}{32}\times \frac{W}{32}\times C$, where $H$, $W$, and $C$ are the height, width, and channel dimension. Two simple fully-connected~(FC) layers are followed~(64 hidden nodes) to regress $z$ into quality scores $q$ with the size of $\frac{H}{32}\times \frac{W}{32}\times 1$. The overall quality is obtained by averaging $q$ on the spatial.

As for videos, the size of prepared data is $S=H\times W\times T$, where $T$ is the temporal dimension. Given a $4\times4\times2$ embedding patch, the size of $z$ is $\frac{H}{32}\times \frac{W}{32}\times \frac{T}{2}\times C$. Similarly, we also adopt two FC layers for the regression, and obtain the scores $q$ with the size of $\frac{H}{32}\times \frac{W}{32}\times \frac{T}{2}\times 1$. The global average is adopted to get the overall quality.

\subsection{Relative Scale Encoding}

The only concern is whether the model can distinguish the scale-mixed data. To clarify the concern, we explore various relative scale encoding methods in the model to explicitly highlight the scale information.
For simplicity, we select the video data with the progressive temporal mask for example. 

\subsubsection{SAMA-W} Instead of directly averaging $q$, SAMA-W introduces extra FC layers to learn $\frac{T}{2}$ weights in temporal dimension~(corresponding to different scales), and compute a weighted average for the final quality.

\subsubsection{SAMA-SE} In this variation, the SE module~\cite{se2018} is introduced in the transformer block. The SE module is set after the attention operation but before the residual connection in each transformer block. Still, we only focus on the temporal dimension where the features are squeezed and excited with a regular SE module.

\subsubsection{SAMA-RSB-A} The relative scale bias is explored in the transformer block. In the original SwinTransformer, the relative position bias is embedded in the attention module as
\begin{equation}
Attn(Q, K, V) = \mathrm{SoftMax}(QK^T/\sqrt{d} + B)V, 
\end{equation}
where $Q$, $K$, $V$, $B$, and $d$ are query, key, value, relative position bias, and channel dimension~\cite{swin2021}. In order to tune the scale information adaptively, an additional relative scale bias~(denoted as $R$) is introduced, and the attention is modified as
\begin{equation}
Attn(Q, K, V) = \mathrm{SoftMax}(QK^T/\sqrt{d} + B + R)V.
\end{equation}

\subsubsection{SAMA-RSB-M} The relative scale bias $R$ can be also introduced as a multiplication factor instead of the addition term, which is formulated as 
\begin{equation}
Attn(Q, K, V) = \mathrm{SoftMax}((QK^T/\sqrt{d} + B)\cdot R)V.
\end{equation}

Note that these encoding methods are not included in our basic SAMA method. With SAMA, we just explore some variations in model architecture based on data characteristics, and see if both of them are cooperatively enhanced. 

\begin{table*}[htbp]
  \centering
    \resizebox{0.85\linewidth}{!}{ 
    \begin{tabular}{cccccccccccc}
    \toprule
    \multicolumn{4}{c}{Trained on LSVQ} & \multicolumn{2}{c}{LSVQ} & \multicolumn{2}{c}{LSVQ-1080p} & \multicolumn{2}{c}{KoNViD} & \multicolumn{2}{c}{LIVE-VQC} \\
\cmidrule{1-4}    Type  & Method & Year  & Publication & SRCC  & PLCC  & SRCC  & PLCC  & SRCC  & PLCC  & SRCC  & PLCC \\
    \midrule
    \multirow{2}[2]{*}{hand-crafted} & TLVQM & 2019  & T-IP  & 0.772 & 0.774 & 0.589 & 0.616 & 0.732 & 0.724 & 0.670 & 0.691 \\
          & VIDEVAL & 2021  & T-IP  & 0.795 & 0.783 & 0.545 & 0.554 & 0.751 & 0.741 & 0.630 & 0.640 \\
    \midrule
    \multirow{6}[2]{*}{multi-branch} & PVQ (w/o patch) & 2021 & CVPR & 0.814 & 0.816 & 0.686 & 0.708 & 0.781 & 0.781 &0.747 & 0.776 \\
          & PVQ (w patch) & 2021 & CVPR & 0.827 &0.828 & 0.711 & 0.739 &0.791 &0.795 & 0.770 & 0.807 \\
          & BVQA-2022 & 2022 & T-CSVT & 0.852 & 0.855 &0.771 & 0.782 & 0.834 & 0.837 & 0.816 & 0.824 \\
          & DSD-PRO & 2022 &T-CSVT & 0.875 & 0.875 & 0.765 & 0.793 & 0.865 & 0.856 & \textbf{0.838} & 0.849 \\
          & ZoomVQA & 2023 & CVPRW & 0.886 & 0.879 &0.799 & 0.819 & 0.877 & 0.875 & 0.814 & 0.833 \\
          & DOVER & 2023 & ICCV & \textbf{0.888} & \textbf{0.889} & \textbf{0.795} & \textbf{0.830} & \textbf{0.884} & \textbf{0.883} & 0.832 & \textbf{0.855} \\
    \midrule
    \multirow{8}[8]{*}{single-branch} & VSFA  & 2019  & MM    & 0.801 & 0.796 & 0.675 & 0.704 & 0.784 & 0.795 & 0.734 & 0.772 \\
          & FAST-VQA & 2022  & ECCV  & 0.876 & 0.877 & 0.779 & 0.814 & 0.859 & 0.855 & 0.823 & 0.844 \\
\cmidrule{2-12}          & Baseline &       & \textcolor[rgb]{ .5, .5, .5}{} & 0.870 & 0.871 & 0.766 & 0.801 & 0.860 & 0.855 & 0.816 & 0.836 \\
\cmidrule{2-12} 
          & SAMA (c) &       &   & 0.879 & 0.880 & \textbf{0.785} & \textbf{0.825} & 0.879 & 0.877 & 0.816 & 0.839 \\
          & SAMA (c + p) &       &   & 0.880 & 0.881 & 0.783 & 0.821 & 0.879 & 0.877 & 0.821 & 0.838 \\
          & SAMA  &       &   & \textbf{0.883} & \textbf{0.884} & 0.782 & 0.822 & \textbf{0.880} & \textbf{0.877} & \textbf{0.834} & \textbf{0.849} \\
\cmidrule{2-12}          & SAMA (+ spm) &       &  & 0.878 & 0.878 & 0.776 & 0.814 & 0.872 & 0.872 & 0.818 & 0.840 \\
          & SAMA(+ swm) &       &   & 0.877 & 0.878 & 0.780 & 0.820 & 0.873 & 0.867 & 0.815 & 0.836 \\
    \bottomrule
    \end{tabular}%
    }
  \caption{VQA Performance Comparison when Trained on LSVQ (The default SAMA is with progressive temporal mask, while `c'/`c+p'/`spm'/`swm' indicates choppy/choppy+progressive/spatial patch-based/spatial window-based mask, respectively.)}
  \label{tab:vqalsvq}%
  
\end{table*}%

\section{Experiments}

\subsection{Benchmark Datasets and Evaluation Protocols}

We evaluate our proposed method on both VQA and IQA tasks. For VQA task, four datasets~(i.e., LSVQ~\cite{patchvqa2021}, KoNViD~\cite{konvid}, LIVE-VQC~\cite{livevqc} and YouTube-UGC~\cite{youtube}) are adopted to conduct the intra- and inter-dataset validation. For IQA task, we adopt two widely-used datasets~(i.e., KonIQ~\cite{koniq}, and SPAQ~\cite{spaq}). By convention, when trained on LSVQ, we report the performance results on both LSVQ test set, LSVQ 1080p set, KoNViD, and LIVE-VQC. When conducting the intra-dataset validation for both VQA and IQA, the experiment is repeated for 10 rounds and the median performance is adopted to alleviate the randomness.
 To quantify the performance of different methods, Spearman rank-order correlation coefficient~(SRCC) and Pearson linear correlation
coefficient~(PLCC) are adopted for the criteria. A better model would result in higher SRCC and PLCC values.

\subsection{Experiments on VQA}

Our proposed sampling method is verified with 9 newly released VQA algorithms, which are TLVQM~\cite{tlvqm2019}, VIDEVAL~\cite{videval2021}, VSFA~\cite{vsfa2019}, Patch-VQ~\cite{patchvqa2021}, BVQA-2022~\cite{bvqa2022}, DSD-PRO~\cite{dsdpro2022}, FAST-VQA~\cite{fastvqa2022}, ZoomVQA~\cite{zoomvqa2023}, and DOVER~\cite{dover2023}). 
Following the setting in FAST-VQA, models are first trained on LSVQ training set, and the performance is validated on LSVQ test set, LSVQ 1080p set, KoNViD, and LIVE-VQC. 
The performances of existing methods are directly from the published papers, and the comparison is shown in Tab.~\ref{tab:vqalsvq}. We reproduce FAST-VQA as the baseline model in our environment. The default SAMA model is built on the progressive temporal masking. We denote the one with choppy temporal mask as SAMA(c), and the one with mixed mask as SAMA(c+p). We also introduce two more with spatiotemporal masks, where `spm'/`swm' means spatial patch-based/window-based mask respectively.

\begin{figure}[t]  
\centering
\setlength{\abovecaptionskip}{0.2cm}
\subfloat[Train loss]{\includegraphics[width=.46\linewidth]{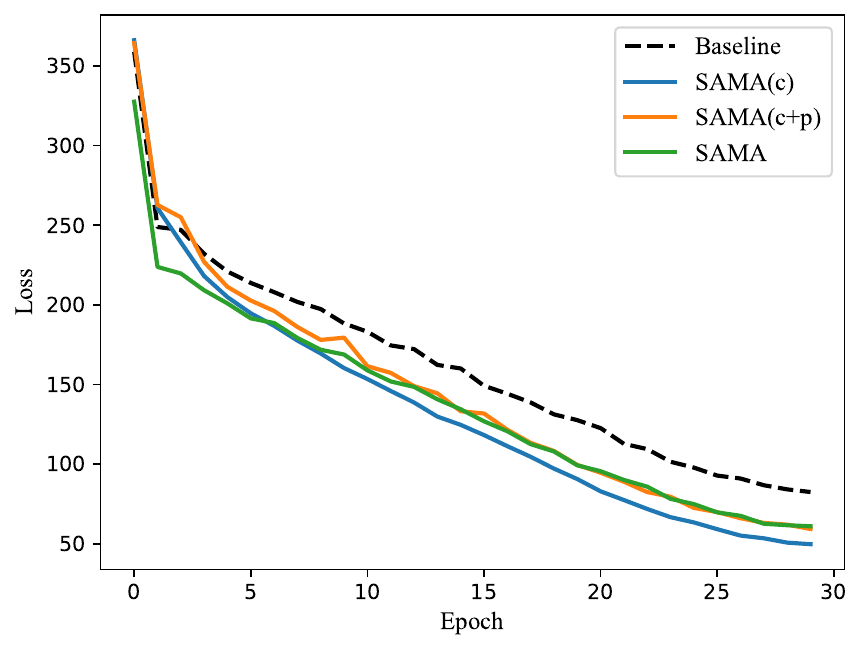}}\quad
\subfloat[Test SRCC]{\includegraphics[width=.46\linewidth]{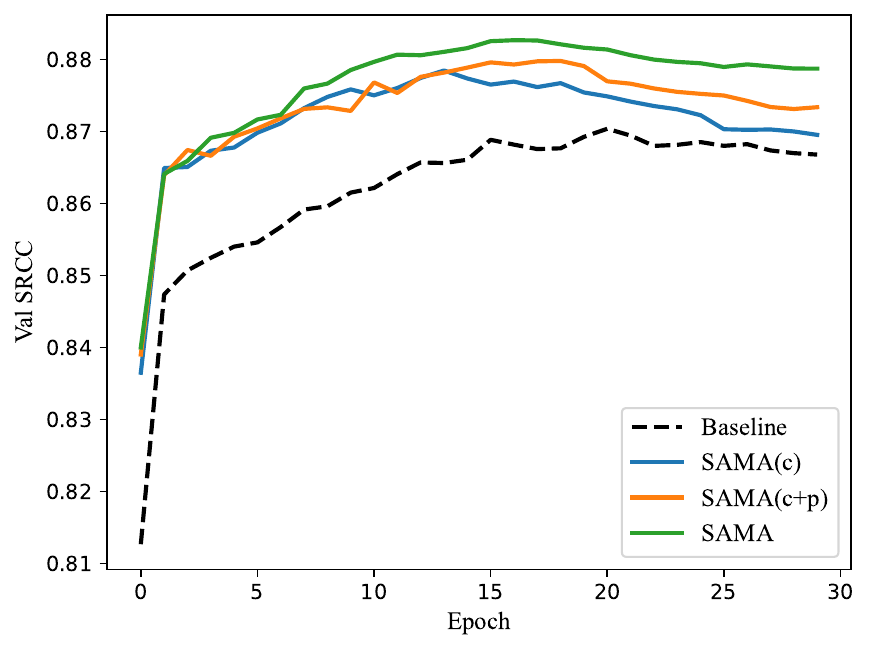}}
\caption{The illustration of training and testing on LSVQ}

\label{fig:trainloss}

\end{figure}

As given in Tab.~\ref{tab:vqalsvq}, with the proposed sampling, our models surpass the baseline model and achieve good results. Since the only difference between the baseline and our models is the proposed sampling method, the improvement demonstrates the effectiveness of our proposed SAMA method. Further, Fig.~\ref{fig:trainloss} gives the training loss and the testing SRCC on LSVQ dataset. With SAMA, our models reach lower training errors, and higher testing SRCC. It is also observed in Tab.~\ref{tab:vqalsvq} that the mixture of spatial and temporal masks may decrease the performance, which is reasonable because the complex pattern might make it harder to learn a good representation, especially in the small-scale dataset. In the table, the best performances in single-branch and multi-branch architectures are highlighted in bold. Note that we weaken the multi-branch models, because our work is focusing the sampling method. Even so, with SAMA, our single-branch models still achieve similar performance to the recent multi-branch models~(mixture of sampling or dual-stream decomposition).

Moreover, some relative scale encoding methods are also explored in Tab.~\ref{tab:rse}. From the table, it shows that the additional encoding modules in the architecture do not help much, with little performance improvement. The result may also suggest that the input data from SAMA can be well recognized and inferred by the model architecture, thus additional relative scale encoding modules are redundant. This result is helpful. It indicates that we do not need to change any of the model, and only a simple replacement of the data sampling can improve the performance.

The intra-dataset performance comparison of VQA is also conducted on LIVE-VQC, KoNViD, and YouTube-UGC. As shown in Tab.~\ref{tab:vqafinetune}, our default model greatly enhances the baseline model~(i.e., the reproduced FAST-VQA). The proposed method is based on the single-branch model, and achieves competitive performance to multi-branch methods with the change of sampling only.

\begin{table*}[t]
  \centering
    \begin{tabular}{ccccccccc}
    \toprule
    & \multicolumn{2}{c}{LSVQ} & \multicolumn{2}{c}{LSVQ-1080p} & \multicolumn{2}{c}{KoNViD} & \multicolumn{2}{c}{LIVE-VQC} \\
    Method  & SRCC  & PLCC  & SRCC  & PLCC  & SRCC  & PLCC  & SRCC  & PLCC \\
    \midrule
    
	Baseline  & 0.870 & 0.871 & 0.766 & 0.801 & 0.860 & 0.855 & 0.816 & 0.836 \\
       SAMA  &  0.883 & 0.884 & 0.782 & 0.822 &0.880 & 0.877 & 0.834 & 0.849 \\
	SAMA-W & 0.882 & 0.883 & 0.783 & 0.822 &0.880 & 0.874 & 0.833 & 0.851 \\
	SAMA-SE & 0.881& 0.882 & 0.785 &0.824 & 0.879 & 0.878 & 0.834 & 0.846 \\
	SAMA-RSB-A &0.881 &0.883 &0.783 &0.823 &0.877 &0.875 & 0.834 & 0.849 \\
	SAMA-RSB-M &0.883 & 0.884 &0.787 &0.825 &0.879 &0.875 &0.835 & 0.848 \\
    \bottomrule
    \end{tabular}%
  \caption{VQA Performance with Relative Scale Encoding}
  \label{tab:rse}%
  
\end{table*}%

\begin{table}[t]
  \centering
    \resizebox{0.98\linewidth}{!}{ 
    \begin{tabular}{ccccccc}
    \toprule
    \multirow{2}[2]{*}{Method} & \multicolumn{2}{c}{LIVE-VQC} & \multicolumn{2}{c}{KoNViD} & \multicolumn{2}{c}{YouTube-UGC} \\
          & SRCC  & PLCC  & SRCC  & PLCC  & SRCC  & PLCC \\
    \midrule
    TLVQM & 0.799 & 0.803 & 0.773 & 0.768 & 0.669 & 0.659 \\
    VIDEVAL & 0.752 & 0.751 & 0.783 & 0.780 & 0.779 & 0.773 \\
    \midrule
    \textcolor[rgb]{ .502,  .502,  .502}{PVQ} & \textcolor[rgb]{ .502,  .502,  .502}{0.827} & \textcolor[rgb]{ .502,  .502,  .502}{0.837} & \textcolor[rgb]{ .502,  .502,  .502}{0.791} & \textcolor[rgb]{ .502,  .502,  .502}{0.786} & \textcolor[rgb]{ .502,  .502,  .502}{-} & \textcolor[rgb]{ .502,  .502,  .502}{-} \\
    \textcolor[rgb]{ .502,  .502,  .502}{BVQA-2022} & \textcolor[rgb]{ .502,  .502,  .502}{0.834} & \textcolor[rgb]{ .502,  .502,  .502}{0.842} & \textcolor[rgb]{ .502,  .502,  .502}{0.834} & \textcolor[rgb]{ .502,  .502,  .502}{0.836} & \textcolor[rgb]{ .502,  .502,  .502}{0.818} & \textcolor[rgb]{ .502,  .502,  .502}{0.826} \\
    \textcolor[rgb]{ .502,  .502,  .502}{DSD-PRO} & \textcolor[rgb]{ .502,  .502,  .502}{\textbf{0.875}} & \textcolor[rgb]{ .502,  .502,  .502}{0.862} & \textcolor[rgb]{ .502,  .502,  .502}{0.861} & \textcolor[rgb]{ .502,  .502,  .502}{0.861} & \textcolor[rgb]{ .502,  .502,  .502}{0.841} & \textcolor[rgb]{ .502,  .502,  .502}{0.836} \\
    \textcolor[rgb]{ .502,  .502,  .502}{DOVER} & \textcolor[rgb]{ .502,  .502,  .502}{0.860} & \textcolor[rgb]{ .502,  .502,  .502}{\textbf{0.875}} & \textcolor[rgb]{ .502,  .502,  .502}{\textbf{0.909}} & \textcolor[rgb]{ .502,  .502,  .502}{\textbf{0.906}} & \textcolor[rgb]{ .502,  .502,  .502}{\textbf{0.890}} & \textcolor[rgb]{ .502,  .502,  .502}{\textbf{0.891}} \\
    \midrule
    VSFA  & 0.773 & 0.795 & 0.773 & 0.775 & 0.724 & 0.743 \\
    FAST-VQA & 0.849 & 0.862 & 0.891 & 0.892 & 0.855 & 0.852 \\
    \midrule
    Baseline & 0.847 & 0.867 & 0.881 & 0.880 & 0.842 & 0.840 \\
    SAMA  & \textbf{0.860} & \textbf{0.878} & \textbf{0.892} & \textbf{0.892} & \textbf{0.881} & \textbf{0.880} \\
    \bottomrule
    \end{tabular}%
    }
  \caption{VQA Performance within Each Dataset}
  \label{tab:vqafinetune}%
  
\end{table}%

\begin{table}[t]
  \centering
    \resizebox{0.98\linewidth}{!}{ 
    \begin{tabular}{ccccccc}
    \toprule
    \multirow{2}[2]{*}{Method} & \multirow{2}[2]{*}{Year} & \multirow{2}[2]{*}{Publication} & \multicolumn{2}{c}{KonIQ} & \multicolumn{2}{c}{SPAQ} \\
          &       &       & SRCC  & PLCC  & SRCC  & PLCC \\
    \midrule
    HyperIQA & 2020  & CVPR  & 0.906  & 0.917  & 0.916  & 0.919  \\
    MUSIQ & 2021  & ICCV  & 0.916  & 0.928  & 0.917  & 0.921  \\
    IEIT  & 2022  & T-CSVT & 0.892  & 0.916  & 0.917  & 0.921  \\
    CVC-IQA & 2022  & T-MM  & 0.915  & 0.941  & \textbf{0.925 } & \textbf{0.929 } \\
    VT-IQA & 2022  & T-CSVT & 0.924  & 0.938  & 0.920  & 0.925  \\
    DACNN & 2022  & T-CSVT & 0.901  & 0.912  & 0.915  & 0.921  \\
    SARQUE & 2022  & T-MM  & 0.901  & 0.923  & 0.918  & 0.922  \\
    CONTRIQUE & 2022  & T-IP  & 0.894  & 0.906  & 0.914  & 0.919  \\
    VCRNet & 2022  & T-IP  & 0.894  & 0.909  & -     & - \\
    DEIQT  & 2023  & AAAI  & 0.921  & 0.934  & 0.919  & 0.923  \\
    Re-IQA & 2023  & CVPR  & 0.923  & 0.914  & 0.918  & 0.925  \\
    LIQE  & 2023  & CVPR  & 0.919  & 0.908  & -     & - \\
    QPT   & 2023  & CVPR  & 0.927  & 0.941  & \textbf{0.925 } & 0.928  \\
    \midrule
    Baseline &       &       & 0.923  & 0.934  & 0.908  & 0.911  \\
    SAMA (spm) &       &       & 0.926  & 0.940  & 0.923  & 0.926  \\
    SAMA  &       &       & \textbf{0.930 } & \textbf{0.942 } & \textbf{0.925 } & \textbf{0.929 } \\
    \bottomrule
    \end{tabular}%
    }
  \caption{IQA Performance within Each Dataset}
  \label{tab:iqa}%
  
\end{table}%

 \begin{figure}[t]  
 \centering
\begin{minipage}[b]{\columnwidth}
 
 \subfloat[Raw data]{
 \centering
\begin{minipage}{0.4\columnwidth}
\includegraphics[width=.865\linewidth]{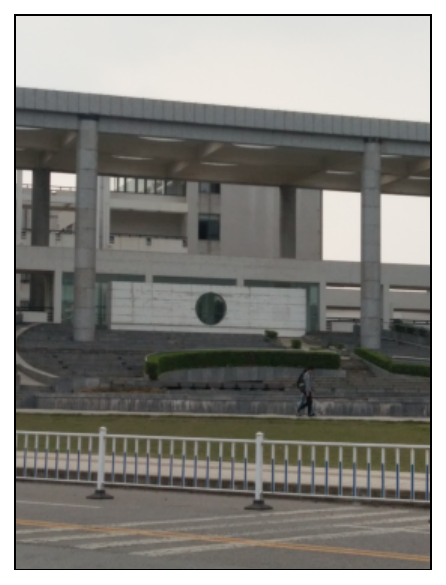}\\
\includegraphics[width=.865\linewidth]{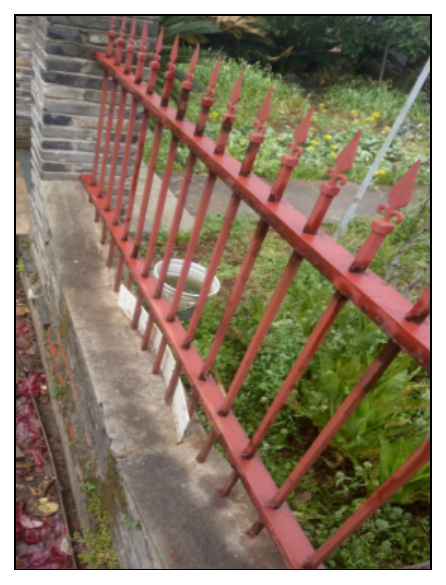}
\end{minipage}
}  
\subfloat[Baseline]{
 \centering
\begin{minipage}{0.28\columnwidth}
\includegraphics[width=.8\linewidth]{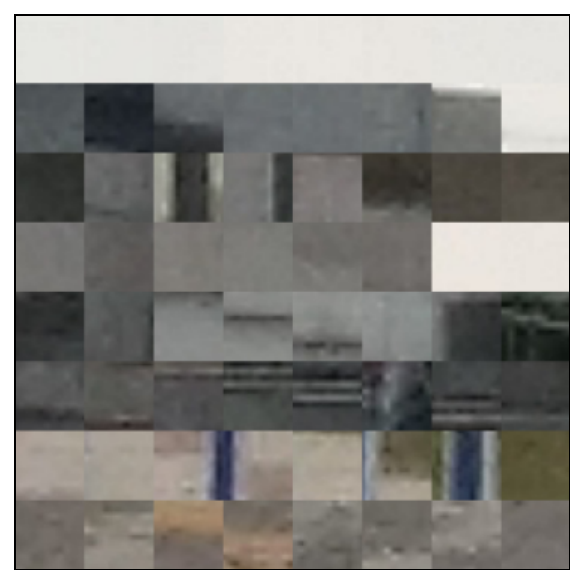}\\
\includegraphics[width=.8\linewidth]{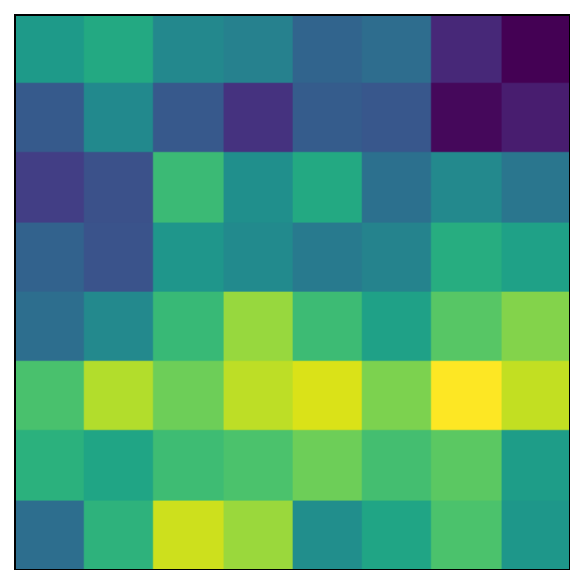} \\
\includegraphics[width=.8\linewidth]{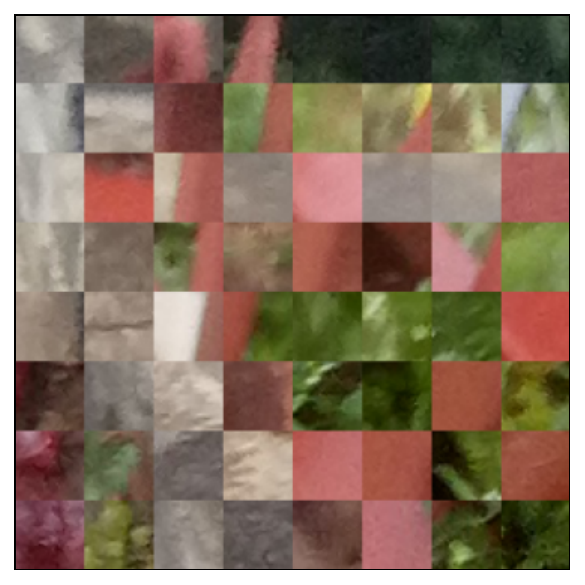} \\
\includegraphics[width=.8\linewidth]{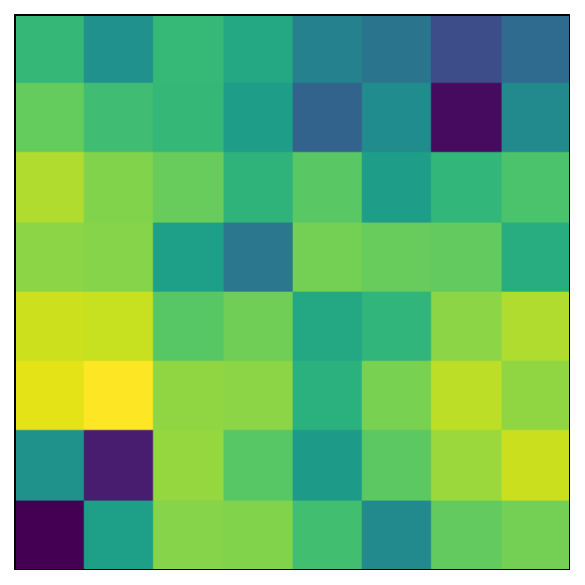}
\end{minipage}
} 
\subfloat[SAMA]{
 \centering
\begin{minipage}{0.28\columnwidth}
\includegraphics[width=.8\linewidth]{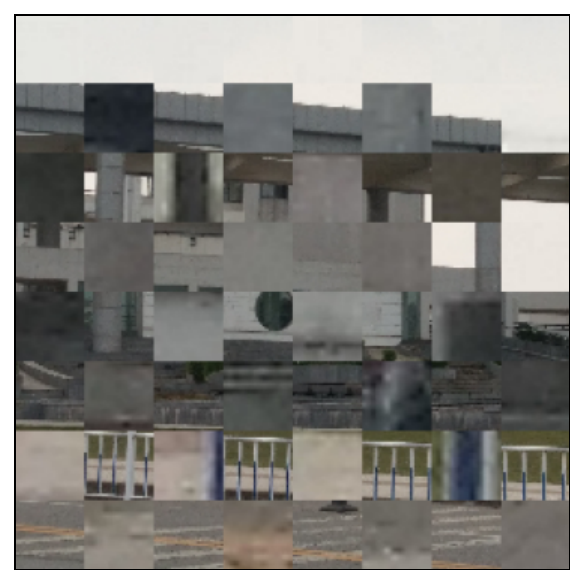}\\
\includegraphics[width=.8\linewidth]{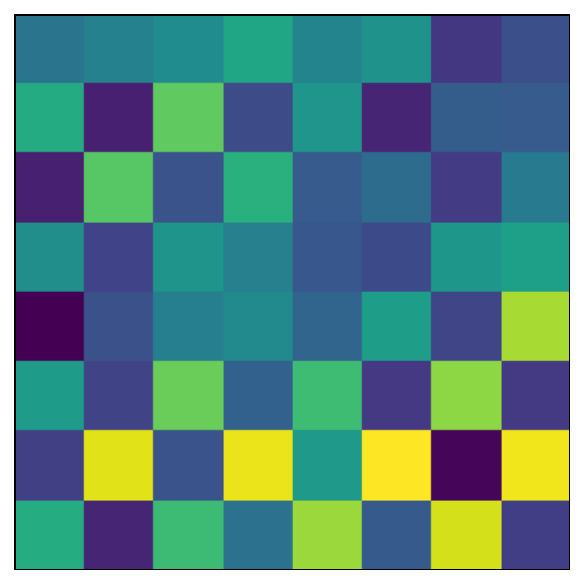} \\
\includegraphics[width=.8\linewidth]{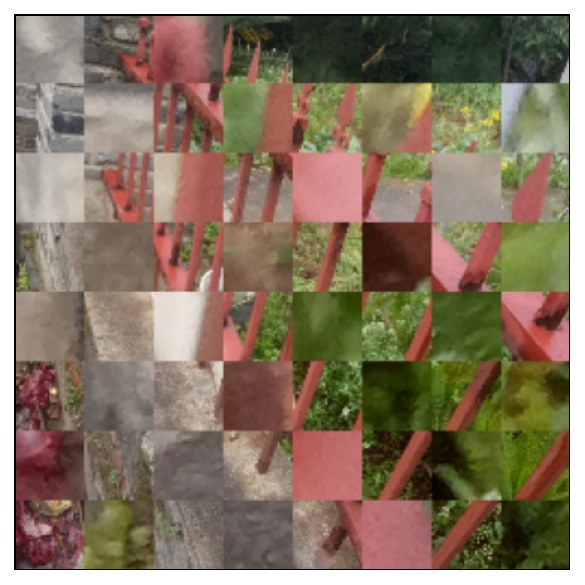} \\
\includegraphics[width=.8\linewidth]{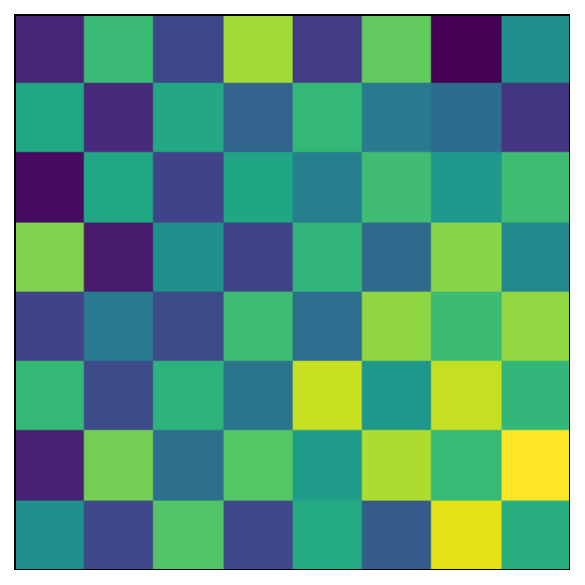}
\end{minipage}
}
\end{minipage}
%
%
\caption{An illustration of fragments~(the first row) and model response~(the second row). }
\label{fig:response}
 
\end{figure}

\subsection{Experiments on IQA}

In this subsection, we verify the proposed method on IQA task. We adopt 13 IQA algorithms in the recent three years for the performance comparison, which are HyperIQA~\cite{hyperiqa2020}, MUSIQ~\cite{musiq2021}, IEIT~\cite{ieit2022}, CVC-IQA~\cite{cvciqa2022}, VT-IQA~\cite{vtiqa2022}, DACNN~\cite{dacnn2022}, SARQUE{sarque2022}, CONTRIQUE~\cite{contrique2022}, VCRNet~\cite{vcrnet2022}, DEIQT~\cite{deiqt2023}, Re-IQA~\cite{reiqa2023}, LIQE~\cite{liqe2023} and QPT~\cite{qpt2023}. Our baseline model is based on the naive SwinTransformer. The spatial window-based mask is adopted as our default SAMA method, and the spatial patch-based mask is denoted as `spm'. 

As shown in Tab.~\ref{tab:iqa}, the proposed sampling method significantly improves the baseline model, and achieves the best performance on the databases. Note that KonIQ and SPAQ are two different datasets, where KonIQ is mainly the public images while SPAQ is specified to smartphone photography. The improvement on both datasets verifies the proposed sampling method can deal with various image content. More importantly, our baseline is simple enough while existing methods might involves specific pretraining~\cite{contrique2022, qpt2023}. The simplicity of SAMA and its promising performance demonstrate the effectiveness of our method. Fig.~\ref{fig:response} gives an illustration of how the model estimates quality. In the figure, (b) and (e) are the sampled fragments from the baseline~(i.e., grid-based fragment sampling, the left column) and SAMA~(the right column), while (c) and (f) are the output responses~(i.e., the predicted quality scores on each window, the lower the darker). From the figure, it shows that our SAMA enhances the ability of global perception, thus improving the baseline method.

\subsection{Complexity and Transferability}

SAMA is a data sampling method and does not involves any change in model architecture, thus the model complexity is the same as FAST-VQA~\cite{fastvqa2022}. Indeed, SAMA is thought to cost only more computation for the interpolation when scaling the data into a pyramid. And as for the sampling and masking procedures, the implementation can be simplified and cost the same consumption as FAST-VQA. This is because both of the methods are to sample the same amount of fragments. The only difference is where the data is sampled from  -- FAST-VQA samples from the raw data, while SAMA samples from the pyramid. 

Our SAMA method is expected to cover more tasks and more models. In this work, we verify the method in two resolution-related tasks~(i.e., IQA and VQA). Actually, the method should be task-agnostic. These tasks benefiting from multi-scale features are also thought to be the applications of SAMA, and we are expecting to extend SAMA to more tasks. Further, We would also try more model architectures~(not only SwinTransformer). The data sampled from SAMA is expected to suit most transformer-based models, which will be examined in our future work.

\section{Conclusion}

Various tasks are benefiting from the multi-granularity representation to bold both the local details and the global semantics, but current data sampling methods cannot balance the local and global paradox simultaneously. The common solution is to increase the input dimension, and thus burden the model complexity. In this work, we have explored a more elegant data sampling method, and put the multi-granularity representation in a regular size with a masking strategy. Experiments on both IQA and VQA have shown that the proposed method is effective with only a single-branch structure. We have also explored various relative scale encoding methods and the results have demonstrated that the proposed SAMA has a good compatibility with the model. We are expecting to apply our method to more basic tasks.

\section{Acknowledgments}
This work was supported by Postdoctoral Research Grant of Shaanxi Province (2023BSHEDZZ165).

\bibliography{aaai24}

\appendix

\section{Implementation Details}
Our basic environment is with Python 3.8.10, Pytorch 1.7.0 and CUDA 11.0 on Ubuntu 18.04. We hold a single NVIDIA 3090 GPU~(24GB) for all the experiments. And the detailed implementation for image quality assessment~(IQA) and video quality assessment~(VQA) is given below.

\subsection{Video Quality Assessment}

The implementation for VQA is based on FAST-VQA~\cite{fastvqa2022}, and the only different is we substitute the original fragment smapling with our proposed SAMA. The `Baseline' in all the experiments of VQA means the reproduced FAST-VQA in our environment. The VideoSwin~\cite{videoswin2022} pretrained on Kinetics400 is adopted as the backbone. The base learning rate is set to 1e-4 for the backbone and 1e-3 for the regression head. The model is trained for 30 epochs. A linear warm-up of learning rate is set for the first 2.5 epochs and the cosine annealing follows afterwards. The batch size is set to 12 as the original 16~(in FAST-VQA) would cause out of memory in our experiments. A weighted average of PLCC loss and ranking loss is adopted, and we optimize the model with AdamW. 
We sample $7\times7$ grids for a training video sample, and in each grid, a $H\times W\times T = 32\times32\times32$ fragment is sampled. Thus, the input dimension is $224\times224\times32$. The embedding patch is set to $4\times4\times2$. In the inference, 128 frames are sampled from a testing video and divided into 4 snippets. The final quality is the average.

\subsection{Image Quality Assessment}

We modified the implementation of FAST-VQA and construct a baseline model for IQA. We adopt the newly released SwinTransformer~\cite{swinv22022} pretrained on ImageNet as the backbone. The input size is $256\times256$. Thus, we sample $8\times8$ grids and in each grid a $32\times32$ patch is sampled. The baseline model samples data with fragments while the proposed model does with SAMA. The learning rate is set to 1e-4 for the backbone, and 1e-3 for the regression head. We train the model for 50 epochs and a linear warm-up is set for the first 5 epochs. The batch size is set to 64. The smooth $L_1$ loss and AdamW are adopted for optimization.

\section{Relative Scale Encoding}
 
\begin{figure}[]  
\centering
\setlength{\abovecaptionskip}{0.15cm}
\subfloat[SAMA-W]{\includegraphics[width=.7\linewidth]{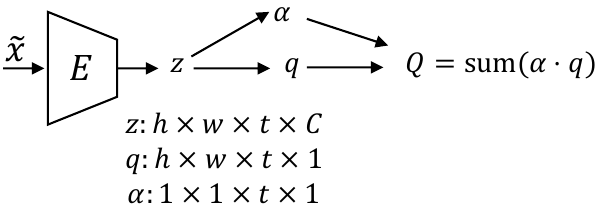}}\\
\subfloat[SAMA-SE]{\includegraphics[width=.7\linewidth]{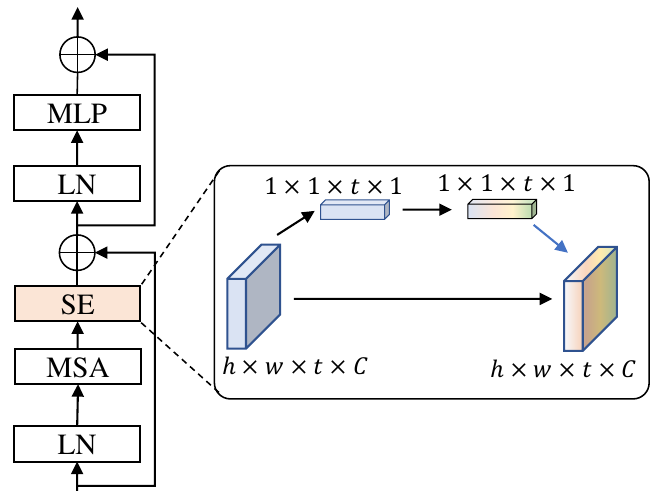}}
\caption{The illlustration of SAMA-W and SAMA-SE.}
\label{fig:rse}
\end{figure}

 We also provide the structure illustration for SAMA-W and SAMA-SE in relative scale encoding~(RSE) of the manuscript. The structure is shown in Fig.~\ref{fig:rse}. Note that RSE is based on our VQA task and the progressive temporal mask is adopted. In this setting, the time dimension corresponds to the scale dimension. Thus, the weighting scheme and the SE module~\cite{se2018} are designed along the time dimension.  

\end{document}